\documentclass{article}
\usepackage{ijcai24}

\usepackage{times}
\usepackage{soul}
\usepackage{url}
\usepackage[hidelinks]{hyperref}
\usepackage[utf8]{inputenc}
\usepackage[small]{caption}
\usepackage{graphicx}
\usepackage{amsmath}
\usepackage{amsthm}
\usepackage{booktabs}
\usepackage{algorithm}
\usepackage{algorithmic}
\usepackage[switch]{lineno}
\usepackage{multirow}

\title{

On the Resurgence of Recurrent Models for Long Sequences:\\  Survey and Research Opportunities in the Transformer Era
}

\author{
Matteo Tiezzi$^1$\and
Michele Casoni$^1$\and 
Alessandro Betti$^2$\and \\
Tommaso Guidi$^1$\and
Marco Gori$^{1}$\And
Stefano Melacci$^1$
\affiliations
\large
$^1$DIISM, University of Siena, Siena, Italy,$\ \ \ $ 
$^2$Inria, Lab I3S, MAASAI, Universit\`e C\^ote d'Azur, Nice, France\\
\emails
\small
matteo.tiezzi@unisi.it,
m.casoni@student.unisi.it,
alessandro.betti@inria.fr,\\
marco.gori@unisi.it,
t.guidi@student.unisi.it,
mela@diism.unisi.it
}

\usepackage[bbgreekl]{mathbbol}
\usepackage{algorithmic}
\usepackage{algorithm}
\usepackage{array}
\usepackage[caption=false,font=normalsize,labelfont=sf,textfont=sf]{subfig}
\usepackage{textcomp}
\usepackage{stfloats}
\usepackage{url}
\usepackage{booktabs}
\usepackage{verbatim}
\usepackage{graphicx}
\usepackage{graphicx}
\usepackage[linguistics]{forest}

\usepackage{tikz}
\usetikzlibrary{matrix,shapes,arrows,fit,tikzmark}

\usepackage{tikz-dependency}

\usepackage{mathtools}
\newcommand{\defeq}{\vcentcolon=}

\newcommand{\loss}{\mathcal{L}}

\def\R{\mathbb{R}}

\newcommand\paragrafo[1]{\medbreak\noindent\textbf{#1.$\quad$}}

\definecolor{dgreen}{rgb}{0, 0.5, 0}

\newcommand{\din}{{d_{\text{in}}}}
\newcommand{\dout}{{d_{\text{out}}}}
\newcommand{\dstate}{{d_{\text{s}}}}
\newcommand{\dk}{{d_{\text{k}}}}

\newcommand{\dkern}{{d_{\text{ker}}}}
\newcommand{\dseq}{L}

\newcommand{\similarity}[1]{\text{sim}\left(#1\right)}

\begin{document}

\maketitle

\begin{abstract}
A longstanding challenge for the Machine Learning community is the one of developing models that are capable of processing and learning from very long sequences of data. The outstanding results of Transformers-based networks (e.g., Large Language Models) promotes the idea of parallel attention as the key to succeed in such a challenge, obfuscating the role of classic sequential processing of Recurrent Models. However, in the last few years, researchers who were concerned by the quadratic complexity of self-attention have been proposing  a novel wave of neural models, which  gets the best from the two worlds, i.e., Transformers and Recurrent Nets. Meanwhile, Deep Space-State Models emerged as robust approaches to function approximation over time, thus opening a new perspective in learning from sequential data, followed by many people in the field and exploited to implement a special class of (linear) Recurrent Neural Networks. This survey is aimed at providing an overview of these trends framed under the unifying umbrella of Recurrence. Moreover, it emphasizes novel research opportunities that become prominent when abandoning the idea of processing long sequences whose length is known-in-advance for the more realistic setting of potentially infinite-length sequences, thus intersecting the field of lifelong-online learning from streamed data.

\end{abstract}

\section{Introduction}
\label{sec:introduction}

\paragrafo{Motivations} The outstanding results achieved by Large Language Models (LLMs) 
and by their even more recent multi-modal variants \cite{geminiteam2023gemini}, rely on attention-based neural architectures with several analogies to the encoder or decoder from the originally proposed Transformers \cite{vaswani2017attention}. Sequential data is processed by attention models, usually handling input tokens in parallel with multi-headed self-attention blocks. The {\it ``attention is all you need''} message   somewhat hindered further developments on the already-established sequential processing of recurrent models \cite{yu2019review,gori_mori1992,goriu_hammer2010}. However, the need of reducing the computational burden and of improving LLMs in both training and inference, has been renewing the interest in recurrent architectures for processing sequential data \cite{orvieto2023resurrecting,peng2023rwkv}. Moreover, seminal approaches in continuous-time Recurrent Neural Networks (RNNs) \cite{voelker2019,gu2020hippo} opened to an alternative path in handling long-range sequences with the so-called State-Space Models (SSMs) \cite{gu2021efficiently}. This path inspired a plethora of works aiming at injecting pure SSMs into deep architectures \cite{gu2021combining,smith2022simplified},  
contributing to the proposal of a new class of powerful recurrent units, Linear Recurrent Units (LRUs) \cite{orvieto2023resurrecting}.

\paragrafo{Contributions} These novel routes, paired with further research in the optimization schemes of RNNs \cite{marschall2020unified,javed2023online}, yielded a new chapter in the narrative of sequence processing  (Section~\ref{sec:keyconcepts}), rooted on the resurgence of recurrence, which is the main topic of this survey (Section~\ref{sec:resurgence}). On one hand, we provide a short review of the main advancements in the field, connecting them with earlier research activities on RNN models, thus allowing a wide range of audience to quickly get into this topic. On the other hand, this paper highlights gaps and research directions which feature interesting challenges, especially in the context of lifelong online learning from sequential data continuously streamed from one or multiple sources, which intrinsically suggests the need of handing potentially infinte-length sequences  (Section~\ref{sec:gapsanddirections}). Such a context represents an inspiring direction for developing novel technologies that go beyond the current massively-offline-trained models based on datasets of finite-length sequences. Finally, we briefly summarize the reviewed findings (Section~\ref{sec:discussion}).

\section{Recurrent Models: Key Concepts}
\label{sec:keyconcepts}
\paragraph{Recurrent models.} In many real-world applications, such as those related to natural language, video processing, time series forecasting, and others, leveraging the order in which patterns are provided to the machine is crucial for discovering cues in the task at-hand \cite{elman1990finding}. Let us consider a sequence of patterns $(u_t)_{t=1}^{\dseq} := (u_1,\dots, u_{\dseq})$, where $\dseq$ is an integer that represents the sequence length and $u_t \in \R^{\din}$ is the $t$-th element of the sequence. 
Recurrent models process one pattern at-a-time, following the sequence order, progressively (sequentially) encoding the information into $\dstate$ context units whose output values model a {\it state} $x_t \in \R^{\dstate}$.
While in standard feed-forward networks hidden neurons develop internal representations of the current input patterns, in recurrent models the state encodes the temporal properties of the input data that was sequentially provided so far.
The state is further processed for generating an {\it output} signal $y_t \in \R^{\dout}$, $\dout \geq 1$. The Elman's architecture \cite{elman1990finding} was one the pillars to the foundations of classic RNNs, and it is based on the following equations, 
\begin{equation}
    x_{t} = \sigma(Ax_{t-1}+ Bu_t), \quad y_t = \sigma_{\text{out}}(C x_t + D u_t), 
    \label{eq:RNN}
\end{equation}
where we considered a \textit{state matrix} $A\in\R^{\dstate \times \dstate}$, an \textit{input matrix} $B\in\R^{\dstate\times \din}$, an \textit{output matrix} $C\in\R^{\dout \times \dstate}$ and a \textit{feedthrough matrix} $D\in\R^{\dout \times \din}$.\footnote{$D\ne 0$ introduces a skip connection that was not included in the original Elman network \cite{elman1990finding}.} We denote with $\sigma$ and $\sigma_{\text{out}}$ non-linearities on the state and output computations, respectively, often selected to be a $\tanh$ or sigmoid activations. If $\sigma$ is the identity function, then we say the RNN model is \textit{linear}. 
When 
$\sigma_{\text{out}}$ is the identity function, linear RNNs are said to be in {\it state-space model form} \cite{schafer2006recurrent}. 
RNNs are able to simulate Universal Turing Machines \cite{siegelmann2012neural}, and RNNs in state-space model form have been proven to be universal approximators \cite{schafer2006recurrent}, i.e., they are able to approximate any open dynamical system with an arbitrary accuracy \cite{li2022approximation}. 
There are several books, 
surveys and reviews  in which the reader can find further details and properties of RNN-based modes \cite{lipton2015critical,salehinejad2017recent,yu2019review}.

\paragrafo{Learning} Let us introduce the instantaneous loss function $\ell(y_k, \hat{y}_k)$ which quantifies to what degree the predicted output $y_k$ matches some supervised/target output $\hat{y}_k$.
Let us collect the model parameters in $\theta := \{A, B, C, D\}$.  
Backpropagation Through Time (BPTT) \cite{rumelhart1985learning} is the de-facto standard algorithm to learn $\theta$ by gradient descent, where the chain-rule is applied over the unfolded network in order to compute gradients \cite{werbos1990backpropagation}. Given a sequence and the empirical risk $\loss(\theta) = \frac{1}{\dseq} \sum_{k=1}^{\dseq} \ell \left( y_{k}, \hat{y}_{k} \right)$, it can be shown that the gradient w.r.t. $\theta$ is a sum of products of partial gradients,
\begin{equation}
\begin{aligned}
\frac{\partial  \loss}{\partial \theta} &= 
\frac{1}{\dseq}   \sum_{k=1}^{\dseq} \frac{\partial \ell(y_{k}, \hat{y}_{k})}{\partial {y}_{k}} \frac{\partial {y}_{k}}{\partial x_{k}} \sum_{j=1}^{k} {\Big( \prod_{s=j}^k \frac{\partial x_{s}}{\partial x_{s-1}}\Big) \frac{\partial x_{j-1}}{\partial \theta}},
\label{eq:bptt}
\end{aligned}
\end{equation}
where a central role is played by the term ${\prod_{s=j}^k {\partial x_{s}}/{\partial x_{s-1}}}$, which \textit{transports} the error back in time from step $k$ to step $j$. We can rewrite this term in the form of a product of Jacobi matrices \cite{pascanu2013difficulty},
\begin{equation}
\begin{aligned}
{\prod_{s=j}^k \frac{\partial x_{s}}{\partial x_{s-1}} = \prod_{s=j}^k A^T \text{diag}\big(\sigma'(x_{s-1})\big)},
\label{eq:product}
\end{aligned}
\end{equation}
where $\text{diag}(\cdot)$ converts a vector into a diagonal matrix, and $\sigma'$ computes the derivative of the activation function $\sigma$ in Eq. \eqref{eq:RNN} in an element-wise fashion.

\paragrafo{Limits} Unless the partial terms ${\partial x_{s}/ \partial x_{s-1}}$ are close to 1, the product in Eq. \eqref{eq:product} could explode or vanish. 
In details, in the simplified case of a linear model (i.e., setting $\sigma$  to the identity function) a sufficient condition for long-term components to vanish is that of having the largest eigenvalue of the state matrix $A$ smaller than one, i.e., $\lambda < 1$. Conversely,  $\lambda > 1$ is a necessary condition for gradients to explode, causing older states to have little contribution. See \cite{pascanu2013difficulty} for further details. When gradients undergo vanishing during their backward propagation through the network, the critical \textit{credit assignment} of backpropagation is compromised. In fact, information regarding minor state changes in the distant past loses its ability to influence future states. Conversely, when gradients explode, optimization algorithms encounter substantial difficulties in traversing the cost surface. 
In order to make RNNs better suited to handle longer sequences, popular variants were proposed and widely used in the scientific literature and several real-world applications, such as Long Short-Term Memories (LSTMs), 
Gated Recurrent Units (GRUs), 
and others \cite{yu2019review}.

\paragrafo{Recurrence vs. attention} While the sum-product of $\dseq$ terms in Eq. \eqref{eq:bptt} results in  $O(\dseq^2)$ computational complexity for each sequence, efficient gradient propagation schemes have been devised to make cost linear \cite{kag2021training,gruslys2016memory}. From a memory standpoint, BPTT requires to store all the intermediate states over time, resulting in an $O(\dseq)$ overhead. These interesting properties, paired with outstanding experimental results, favoured the growth of popularity of RNNs (different variants) within the Deep Learning wave, in language, vision, and beyond \cite{sutskever2014sequence,xu2015show}. However, in 2017, RNNs were shadowed by the Transformers tsunami \cite{vaswani2017attention}. In fact, the authors of \cite{vaswani2017attention} promoted a {\it stateless} architecture, fully based on attention, {\it ``dispensing with recurrence and convolutions entirely''}, where the essence\footnote{Transformers usually include multiple multi-head attention-based layers interleaved by FFNs. We considered a simple head, with no further projections, to better connect it to the RNN case.} of the attention-based encoding process (which in RNNs is Eq.~\eqref{eq:RNN}) becomes 
\begin{equation}
   y_t = \sum_{i=1}^\dseq \frac{ \similarity{q_t, k_i} }
                {\sum_{j=1}^\dseq \similarity{q_t, k_j}} v_i^T,\quad \begin{array}{ll} q_z & \hskip -3mm \defeq q(u_z, W_q)\\ k_z & \hskip -3mm \defeq k(u_z, W_k) \\ v_z & \hskip -3mm \defeq v(u_z, W_v) \end{array}
    \label{eq:general-attn}
\end{equation}
being $\similarity{}$ a similarity function, such as $\similarity{a, b} = \exp{({a^T b}/{\sqrt{|a|}})}$, and $q$, $k$, $v$ are three projections depending on learnable matrices $W_{q}$, $W_{k}$, $W_{v}$, respectively. The dependence on some previous state $x_{t-1}$, as in Eq.~\eqref{eq:RNN}, is not there anymore, and the cost of computing $y_t$ for $\dseq$ steps is quadratic, $O(\dseq^2)$.
Conversely, storing a progressively updated state allows recurrent schemes to achieve $O(1)$ complexity per time step. 
However, in Transformers, each $y_t$ can be computed in parallel to the other ones, 
and this became an attractive property for GPU-oriented implementations. 

\section{Resurgence of Recurrence}
\label{sec:resurgence}
When dealing with long sequences, ($i.$) the quadratic complexity of self-attention in Transformers becomes a serious concern, fostering the need of novel paths to achieve efficient processing \cite{tayefficient}. Moreover, ($ii.$) the need to more accurately preserve extremely long-term dependencies on sequences favoured the proposal of new models. These two requirements ($i.$ and $ii.$) were jointly satisfied by focusing on recurrent formulations, leading to the current resurgent of recurrence \cite{katharopoulos2020transformers,orvieto2023resurrecting}.

\subsection{Attention Marries Recurrence}
\label{sec:tranformers}
Transformers have the capability of capturing dependencies among all the possible input token pairs belonging to the sequence, both local and long-range. In order to reduce the quadratic complexity of {\it stateless} self-attention, many solutions based on {\it stateful} recurrent computations have been introduced, revising the originally proposed attention procedure \cite{katharopoulos2020transformers,peng2023rwkv,sun2023retentive}. We softly subdivide the plethora of scientific papers which follow this research direction as a function of their most evident property.

\paragrafo{Linear transformers} The most evident architectural step toward in reintroducing recurrence is the one of Linear Transformers \cite{katharopoulos2020transformers}, reaching linear complexity (w.r.t. $L$) in self-attention. This is achieved by redefining the similarity function of Eq.~\eqref{eq:general-attn} as a kernel $\mathcal{K}$, $\text{sim}(q, k) := \mathcal{K}(a, b) =  \phi(a)^T \phi(b)$, where $\phi: \R^\dk \mapsto \R^\dkern$ is a non-linear feature map and the kernel codomain is possibly positive in order to define proper attention weights, i.e., $\mathcal{K}: \R^\dk \times \R^\dk \mapsto \R^+$.  
\begin{equation}
\nonumber
y_t = \sum_{i=1}^\dseq \frac{\phi(q_t)^T \phi(k_i) v_i^T}{\sum_{j=1}^\dseq \phi(q_t)^T \phi(k_j)} = \frac{\phi(q_t)^T S_\dseq}{\phi(q_t)^T z_\dseq},
\label{eq:linearatt}
\end{equation}
where matrix $S_\dseq := \sum_{j=1}^\dseq \phi(k_j)  \otimes v_j$, being $\otimes$ the outer product between vectors, and $z_\dseq := \sum_{j=1}^\dseq \phi(k_j)$.
When exploited in an \textit{autoregressive} setting,\footnote{Standard self-attention is not causal, since tokens with index $>t$ can influence the encoding of the $t$-th one. However, autoregressive transformers can be obtained by appropriate masking procedures.} this formulation can be rewritten in a {\it stateful} form, where the internal state $x_t$ is composed by a recurrent state matrix $S_t$ and a recurrent normalization vector $z_t$, i.e., $x_t \defeq (S_t, z_t)$, which are iteratively updated with what is referred to as \textit{additive} interaction,
\begin{eqnarray}
\begin{aligned} x_t &= x_{t-1} + f(u_t) \\ &= \left( \begin{aligned} S_{t-1} &+ \phi(k_t) \otimes v_t\\
 z_{t-1} &+ \phi(k_t) \end{aligned} \right) \end{aligned},\quad\quad\quad  \begin{aligned} y_t &= g(x_t, u_t) \\
 &= \frac{\phi(q_t)^TS_t}{\phi(q_t)^Tz_t} \end{aligned}
\label{eq:rec_transf_out}
\end{eqnarray}
with some initial conditions for $S_0$ and $z_0$ (such as zero) and where $f$ and $g$ are just placeholder functions to emphasize the dependence on data at time $t$, recovering the structure of Eq.~\eqref{eq:RNN}. 
Now $y_t$ can be computed in a recurrent fashion, challenging the traditional distinction between RNNs and Transformers.
If the cost to compute $\phi$ is $O(\dkern)$, then the overall run-time complexity of a Linear Transformer for a sequence of length $\dseq$ is $O{(\dseq \dkern \dk)}$.
The role of the kernel function $\phi$ has been investigated in several subsequent works \cite{peng2021random,schlag2021linear}. Original Linear Transformers exploit $\phi_\text{elu}(x) = \text{elu}(x) + 1$, where $\text{elu}(\cdot)$ denotes the exponential linear unit,  
a kernel that preserves the dimension of the input key vector $(\dkern = \dk)$, yielding $O{(\dseq \dk^2)}$. 

\paragrafo{Improving linear transformers} Despite the effective computational advantages, it turns out that Linear Transformers underperform vanilla self-attention, and the main cause has been attributed to the choice of the kernel function \cite{peng2021random,schlag2021linear}. 
Several works were proposed to overcome this issue, leveraging random feature maps to achieve unbiased estimations of shift-invariant kernels, such as the Gaussian one, in order to approximate the effect of softmax in vanilla attention (see, e.g., 
Random Feature Attention \cite{peng2021random}).
The {TransNormer} model \cite{qin2022devil} identifies the origin of the degradation in performance of kernel-based linear transformer in unbounded gradients and sparse distribution of attention scores. The authors of TransNormer propose to 
exploit a linear kernel and remove the $z_t$ normalization factor in Eq.~\eqref{eq:rec_transf_out} (right), introducing Layer Normalization 
on the attention scores. 
This intuition was already pointed out by
Schlag et al. (DeltaNet \cite{schlag2021linear} and succeeding works that emphasized the formal relationships between kernel-based Linear Transformers and Fast Weight Programmers (FWPs) \cite{schmidhuber1992learning}.
FWPs are built on the intuition of making the model weights input-dependent, with a two-network system where a slow net with slow varying weights continually reprograms a fast net's weights, making them dependent on the spatio-temporal context of a given input stream.
The kernel-based causal perspective of linear attention, Eq.~\eqref{eq:rec_transf_out}, is a FWP with an additional recurrent normalization factor $z_t$. 
Still inspired by connections to FWP, two other ingredients that were studied to improve recurrence are gating functions and the introduction of decay terms, to help modulate the impact of input information.
For instance, recent variants of Linear Transformers such as RetNet, \cite{sun2023retentive}, 
among several other improvements, introduce a forgetting mechanism via a fixed decay factor $\gamma$ in the $S_{t-1}$ term of Eq.~\eqref{eq:rec_transf_out} (left), to control the capability of each token to pool information from its surrounding tokens based on encoding-related priors. 

\paragrafo{Alternative factorization of attention} Kernel-based formulations exploited to linearize self-attention are effective in reducing the computational cost with respect to the sequence length, but they still result in quadratic complexity w.r.t. the feature dimension, $O(\dseq \dk^2)$, which is unfriendly to large model sizes \cite{zhai2021attention}. 
A recent alternative emerged from a low-rank factorization of the $\similarity{}$ function in Eq~\eqref{eq:general-attn}, i.e., $\similarity{a, b}=\sigma_q(a)\psi(b)$. 
Attention Free Transformers  (AFT) \cite{zhai2021attention} implement $\sigma_q$ as an elementwise nonlinear activation function (e.g., sigmoid) and $\psi(a) = e^{a}$, and perform the following operations,
\begin{equation}
y_t = \sigma_q(q_t) \frac{ \sum_{i=1}^{\dseq} e^{k_i + p_{t,i}} v_i}{\sum_{i=1}^{\dseq} e^{k_i+p_{t,i}}} 
\label{eq:aft}
\end{equation}
where $p_{t,i}$ denotes {a learnable position bias involving token $i$ at time $t$}.
This approach eliminates the need of dot product in self-attention by rearranging the order in which computations are performed, and it can still be written in additive recurrent form when dealing with an autoregressive setting, as discussed in the case of Eq.~\eqref{eq:rec_transf_out}. 
Inspired by AFT, the Receptance Weighted Key Value (RWKV) model \cite{peng2023rwkv}  replaces AFT’s position embeddings with an exponential decay, allowing to exploit the model recurrently both for training and inference,  leading to significant gains in efficiency.

\paragrafo{Structuring recurrence} Amidst the multiple advantages brought by the vanilla self-attention mechanism, early attempts to tackle language modeling with Transformers were slacken by the inability ($i.$) to model intra-sequential relations due to the static order dependencies available in standard positional encodings (i.e., the absence of explicit temporal information) and ($ii.$) to propagate inter-sequence information among the processed contexts. Several attempts to tackle these two issues were presented in the last few years. Additionally, such approaches inspired recent sub-quadratic methods that split the overall computation into sequence-chunks which are processed in parallel. Overall, we can distinguish among {\it intra-sequence} recurrent modeling \cite{huang2022encoding}, 
where additional encoders are introduced or embeddings are further structured (split embeddings); {\it segment-level} recurrence, which introduces the ability to temporally
connect different contexts by means of a cache memory (Transformer-XL \cite{Dai2019transformerxl}, Recurrent Memory Transformers \cite{bulatov2022recurrent}); {\it chunck-level} recurrence, in which the input sequence is divided into chunks before processing, opening to sub-quadratic chunk-wise parallel implementations, with (serial) inter-chunk recurrent connections 
 \cite{hutchins2022block}.
Recurrent mechanisms with appropriately structured recurrence have also been proposed in Visual Attention Transformers {(ViT)} 
applied to long video sequences, which are processed sequentially in a sliding window fashion, to keep manageable compute and GPU memory but still ensure relevant information from past windows is within reach. The reader can find details into Section IVB of the thematic survey \cite{selva2023video}.

\subsection{Deep State-space Models as Recurrent Nets}
\label{sec:deepstatespace}
Concurrently with the recurrent-Transformers wave, the need to preserve extremely long-term dependencies in sequences has led to many instances of space-state models (SSMs) combined with deep learning \cite{gu2021efficiently,smith2022simplified}.
SSMs are formalized with Ordinary Differential Equations (ODEs) and, in recent literature, they are commonly implemented considering linear time-invariant systems,
\begin{equation}
    \dot x(t) = Ax(t)+ Bu(t), \quad y(t) = C x(t) + Du(t). 
    \label{eq:ssms}
\end{equation}
While the emphasis on learning with SSMs originated from the experience with RNNs \cite{voelker2019} promoting linear recurrence (Section~\ref{sec:keyconcepts}), deep SSMs followed their own independent path. Recently, they have been formally re-connected to RNNs through a careful neural architecture design \cite{orvieto2023resurrecting}.

\paragrafo{Online function approximation} The development of the line of works behind deep SSMs can be traced back to the
seminal work on Legendre RNNs \cite{voelker2019} and HiPPO \cite{gu2020hippo} in which
the authors propose methods to perform online function  approximation, considering functions of {\it one variable}, defined on the
half line $u\colon [0,+\infty)\to\R$.\footnote{We overloaded the use of symbol $u$ to create a link with the previously introduced concepts.} The problem of finding an approximation of $u$ until the current time $t$, i.e., $u^t:=u|_{I_t}$ 
with $I_t:=(0,t)$, is implemented by:  ($i.$) defining a notion of integrability with appropriate measure; ($ii.$) selecting specific classes of basis functions; ($iii.$) defining {\it online} schemes to update the coefficients of the expansion. 
($i.$) In terms of integrability, HiPPO considers a normalized Lebesgue
measure on $I_t$ (which is the standard choice in $\R^n$). A different choice is explored in Legendre RNNs 
that exploit a measure density which is
constant on a window $[t-\theta,t]$ of size $\theta$ just before the end
point $t$ of the considered temporal instant.
($ii.$)  The class of basis functions with which 
 the approximation is performed 
in the case of HiPPO consists of translated and rescaled Legendre polynomials $v^t_n$. 
A similar choice has been also done
in Legendre RNN.
 Once the basis functions have been selected, the approximation $v^t$ of
the function $u^t$
can be expressed (see~\cite{gu2020hippo}) as a combination involving $N$ coefficients $\{ c_n(t), n=1,\ldots,N \}$ to be determined,
\begin{equation}\nonumber\label{eq:coeff_def}
v^t(t')=\sum_{n=1}^{N} c_n(t) v^t_n(t')\ \ \ \hbox{where}\ \ \ 
v^t_n(t')=\sqrt{2} e_n\Bigl(\frac{2t'}{t}-1\Bigr), 
\end{equation}
being $e_n$ normalized Legendre polynomials,
 $\forall t'\in
[0,t]$.
($iii.$) The just described notions are exploited in the design of a continuous-time time-variant state-space model for the coefficients of the approximation, $\dot c(t) = \tilde{A}(t)c(t)+\tilde{B}(t)u(t)$, being $c(t)$ the vector function collecting all the $c_n(t)$'s, and where a normalization factor $1/t$ is applied to $A$ and $B$ (introducing the dependence on $t$ in $A$ and $B$ and thus yielding $\tilde{A}$ and $\tilde{B}$), as it will be shown shortly.
Computations are appropriately discretized \cite{gu2022parameterization}, so that here we replace function $c(t)$ by vector $c_{t}$,\footnote{We overloaded the notation $t$, using it to indicate both a continuous variable and the index of a discrete step.} and the differential equation is used to devise an update strategy,
\begin{equation}
    c_{t} = \left( 1 - A/t \right)c_{t-1} + t^{-1} B u_{t},
    \label{eq:hippodiscrete}
\end{equation}
where $A$ and $B$ are evaluated by closed forms \cite{gu2020hippo} and not learned from data. 

\paragrafo{Learning in state-space models} In strong analogy with what discussed when introducing Linear Transformers, Eq.~\eqref{eq:hippodiscrete} is a {\it linear recurrence} relation.
In both HiPPO and Legendre RNNs, the authors proposed a hybrid model to apply online function approximation mechanisms to the state of an RNN, in order to help store a compact representation covering the whole past time instants.
In particular, the state of the RNN, $x_t$, is updated not only in function of $x_{t-1}$ and $u_t$, as in Eq.~\eqref{eq:RNN}, but also in function of
$c_{t-1}$, that are the coefficients of the online approximation of a 1-d projection of $x$. 
The leap from this hybrid model to more direct implementations $\dot x(t)=Ax(t)+Bu(t)$ with learnable $A$ and $B$, has been proposed in~\cite{gu2021combining}, in which
single layer SSMs are first introduced and then stacked into deep models, 
while in~\cite{gu2021efficiently} such models are refined to overcome some
computational limitations. 
In these works, the emphasis is on inheriting the experience from the HiPPO theory and to introduce learning once $A$ and $B$  belong to a particular class of structured matrices, exploiting the benefits of linearity in the state update equation (Eq.~\eqref{eq:RNN}).

\paragrafo{Variants of state-space models} The basic principles of deep SSMs have been the subject of several recent works, leading to many variants of SSMs. Notably, S4 \cite{gu2021efficiently}, which exploits specific initialization based on HiPPO theory and learning the state matrix in diagonal form (or  diagonal-low-rank), followed by Diagonal State Spaces, DSSs \cite{gupta2022diagonal}, specifically focussed on the effectiveness of diagonal matrix-based implementations and approximated HiPPO initialization, further theoretically expanded in S4D with focus on diagonal matrices of complex numbers with negative real part \cite{gu2022parameterization}. These works exploit linear recurrence and diagonal state matrices to implement fast computational models. It turns out that the state (resp. output) computation can be rewritten as a convolutional process \cite{massaroli2023laughing}, $x_t = \sum_{z=0}^{t-1}A^{z}Bu_{t-z} = (K_x * u)_t$ (resp. $y_t = \sum_{z=0}^{t-1}(CA^{z}B+\delta_{z} D)u_{t-z} = (K_y * u)_t$, with $\delta_z=1$ only when $z=0$, otherwise $\delta_z=0$), where the power $A^{z}$ is trivial to evaluate due to the diagonal form of $A$. 
Related routes were followed by 
S5 \cite{smith2022simplified} 
introducing parallel scans, multi-input multi-output formulations, 
and other ingredients.
In this case, multiple states at different time instants can be efficiently computed in parallel by associative scans \cite{martin2018parallelizing}. 

\paragrafo{Deep state-space models re-meet recurrent nets} The outstanding results in learning from long sequences achieved by SSMs, paired with efficient training and fast inference, intrigued researchers to better understand the motivations behind their success \cite{orvieto2023resurrecting}. It turned out that some basic ingredients of SSMs which are crucial for their performance can be immediately transposed to the design of RNNs. In particular, referring to Eq.~\eqref{eq:RNN}, they are: ($i.$)  linear recurrence ($\sigma$ replaced by the identity function), ($ii.$) diagonal state matrix $A$ with complex entries and proper initialization (each $A_{jj}$ is a complex number, which is a convenient way to represent the diagonalization of non-symmetric matrices; each $A_{jj}$ is also an eigenvalue of $A$, and it is initialized by sampling the unit disk; in turn $B$ and $C$ become learnable matrices with complex values, and only the real part of $Cx_t$ is preserved when computing $y_t$), ($iii.$) exponential parameterization of the learnable elements in $A$ to favour stability (each $A_{jj}$ is $\exp(-\exp(\nu_j) + i \zeta_j)$, with $\nu_j$ and $\zeta_j$ learnable parameters, where $i$ denotes the imaginary part), ($iv.$) normalization of the $Bu_t$ terms in the state update rule (the $j$-th component of $Bu_t$ is scaled by $\sqrt{1-|A_{jj}|^2}$). These choices are evaluated in \cite{orvieto2023resurrecting}, where the so-called Linear Recurrent Unit (LRU) is proposed. A deep RNN is instantiated by a linear encoder of the input data, followed by a stack of blocks each of them composed by an LRU and an MLP (or a Gated Linear Unit) with same input and output sizes, which introduces non-linearity in the net. Indeed, combining linear RNN layers with nonlinear MLP blocks allows to capture complex mappings without nonlinearities in the recurrence. Overall, these models achieve performances that are similar to the state-of-the art in the Long-Range Arena (LRA) benchmark \cite{tay2020long}.


\subsection{Other Trends in Recurrent Models}
\label{sec:novelrecurrentmodels}
Going beyond Transformers and SSMs, the resurgence of recurrence is witnessed also by the presence of several research activities specifically focused on improving RNNs from different point of views. These activities involve a large variety of approaches, that can be more hardly categorized into well-defined directions and that we only briefly summarize in the following. Some works focus on exploiting diagonal state matrices (as in many SSMs) or, similarly, on formulations fully-based on element-wise products, still keeping non-linear activation functions \cite{li2018independently}. 
Other works try to improve the learning/inference time by exploiting properties of specific classes of recurrent models  or in interleaving recurrence and convolutions \cite{martin2018parallelizing}. 
A large body of literature revisits gating mechanisms, typical of LSTMs. We mention approaches that introduce an additive learnable value to the original forget gate, with the purpose of pushing gate activations away from the saturated regimes \cite{qin2023hierarchically}. A recent trend involves RNN architectures whose processing scheme is inspired by ODEs and dynamical systems \cite{rubanova2019latent}. 
Finally, an orthogonal direction is the one that focuses on learning algorithms for recurrent models (see \cite{marschall2020unified,irie2023exploring} and references therein) which opens to the discussion that is the subject of the next section.


\section{Processing Long Sequences: Gaps and Directions}
\label{sec:gapsanddirections}

\paragrafo{Recent experiences with long sequences} Nowadays, Transformers-based models represent the state-of-the art in several tasks where abundant training data is available, ranging from the domain of natural language (question answering, summarization, classification, $\ldots$) to the one of vision and beyond \cite{geminiteam2023gemini}. 
These models are intrinsically able to process context windows with thousands of tokens. Their recurrent counterparts, e.g., the recent RWKV \cite{peng2023rwkv}, reach a performance that is not far from the one of such models in NLP tasks, without overcoming it yet. In the case of Long Sequence Time-Series Forecasting (LSTF), Transformer-based networks have been recently shown to effectively reach outstanding results (Informer, \cite{informer}), also challenged by more ad-hoc architectures (FSNet, \cite{pham2022learning}).
When it comes to the evaluation of a larger spectrum of long sequence tasks, the Long Range Arena (LRA) benchmark \cite{tay2020long} includes problems with sequences composed of $1,000$ to $16,000$ elements, in the context of natural language, images, mathematics and others. SSMs and LRUs yield state-of-the art performances 
\cite{smith2022simplified,orvieto2023resurrecting,massaroli2023laughing}, overcoming Transformers \cite{zucchet2023online}. 

\paragrafo{Gaps} Two of the most evident directions for future research consist in further improving the quality of SSMs/LRUs, and enabling recurrence-based Transformers to reach and overcome the performance of quadratic-complexity attention. However, there is a major gap that requires more specific studies to go beyond existing technologies, summarized in the following question: what is the transition from processing {\it long} sequences to {\it infinite-length} sequences? In fact, while the ultimate goal of mimicking the human capacity to learn over sequential-data inspired several researchers, all the aforementioned approaches are designed to process datasets of finite-length sequences, mostly learning by BPTT. This is in contrast with the way human learns from perceptual stimuli, which are intrinsically continuous over time and not pre-buffered finite-length sequences randomly shuffled to cope with stochastic gradient descent \cite{collectionlessAI,gori_cal2020}. Moreover, this is also in contrast with the motivations behind seminal works in the field of recurrent models, such as the online learning algorithm of Williams and Peng \cite{williams1990efficient}, and the original formulation of LSTMs \cite{hochreiter1997long}, 
amongst others.\footnote{From \cite{williams1990efficient}:
{\it ``an on-line algorithm, designed to be used to train a network while it runs; no manual state resets or segmentations of the training stream is required''}. From \cite{hochreiter1997long}: LSTMs were introduced with a learning algorithm that unlinke BPTT is {\it ``local in space and time''}, where {\it ``there is no need to store activation values observed during sequence processing in a stack with potentially unlimited size''}. 
}
Nowadays, most the popular research activities in the field of RNNs are not considering the case of learning {\it online} from a continuous, possibly infinite, stream of data (i.e., infinite-length sequence), without resets or segmentations \cite{gori_collas2022}. 


\subsection{Infinite-length Sequences}
\label{sec:opportunities}
Dealing with learning online in a lifelong manner, from a single, continuous stream of sequential data $(u_1, u_2, \ldots, u_{\infty})$ (i.e., data that is not i.i.d., the order of data does matter, having a role in the decision process) is characterized by two main requirements: ($i.$) to compute meaningful gradients at time $t$ that depend on properties of the data processed so far, without unrolling the network over the whole (potentially infinite) input sequence (in the most extreme case only in function of information at time $t$ and $t-1$), updating the model parameters at each $t$; ($ii.$) to have the capability of adapting to the properties of the streamed data, not forgetting the different skills that are gained over time, as commonly stressed in the context of continual learning.
The branch of scientific research that has been working toward the development of online learning algorithms in RNNs, with emphasis on property ($i.$), has been surveyed in \cite{marschall2020unified}. 
Putting aside the heuristic behind Truncated BPTT,  
which, in its naive formulation, consists of
truncating the backward propagation to a fixed number of prior time steps, thus not fully online in strict sense, the most widely known approach to online learning with recurrent models is Real Time Recurrent Learning (RTRL), 
that has been recently revisited to improve its computational and memory requirements \cite{irie2023exploring} 
together with several direct or indirect variants \cite{marschall2020unified}. Some authors recently studied methods to compute gradients in a forward manner exploiting linear recurrence \cite{zucchet2023online}, still depending on a finite-length sequence that is processed without updating the learnable parameters at each $t$, thus not fully satisfying ($i.$).
When considering property ($ii.$), we fall in a setting which can be considered an instance of Online Continual Learning 
with a strong intersection to Streaming Learning, as recently summarized in \cite{streaming}. Notice that this is different from the case of works that focus on continual learning from streams of finite-length sequences. 

\paragrafo{Open challenge} Overall, to our best knowledge, the setting described in this section is largely unexplored. In fact, despite many progresses in the related fields, requirements ($i.$) and ($ii.$) are still already hard to fulfil when faced in an independent manner, thus yielding an even more difficult scenario when joined together. However, it represents the case of learning from the visual data that hit our eyes, the sounds we continuously perceive, the video stream from an always-on camera, language, and so on and so forth. Common benchmarks on sequential data, such as LRA \cite{tay2020long}, are based on collections of long sequences, thus not perfectly suited for the aforementioned setting, while they are good opportunities to evaluate the fulfilment of requirement $(i.)$. It is also practical to work in the typical setting of online learning problems, where all the data is used for training purposes and errors are accumulated while learning (i.e., $u_t$ is processed and $y_t$ is compared with the available ground truth to update the accuracy estimate; then, $y_t$ is  used to evaluate a loss function, compute gradients at $t$ and update the model). LSTF is more flexible from this point of view than classification problems on long sequences, since it offers ground truth at each $t$. Moreover, while the model is not expected to learn new tasks, the properties of the series might change over time thus partially allowing to explore requirement ($ii.$). However, deeply evaluating property ($ii.$) is still beyond what can be concretely done with these datasets. Of course, one might concatenate the sequences taken from offline databases and evaluate a learning problem on such long, single stream, that would be however artificial, due to the intrinsic discontinuity when switching from a sequence to the following one.

\paragrafo{Concrete experience} In order to provide the reader with a concrete experience, we report the results we achieved in LSTF (selected for the aforementioned reasons), using real-world and artificial datasets (see \cite{informer}
 for details), inheriting the same setting of \cite{pham2022learning}, where time-series are split into three disjoint parts, preserving the data order.\footnote{We used/adapted the code of the authors of \cite{informer} and \cite{pham2022learning}, and the library of \cite{marschall2020unified}.} The initial segment of the sequence ($\approx 20\%$) is a training set, followed by a small ($\approx 5\%$) validation set and the remainder of the sequence acts as a test set. These last two parts are still used for online learning, accumulating errors as previously described. Instead of offline-pretraining the model on the training set as in \cite{pham2022learning} and then move to the online processing of the rest of the series, we  process the whole time-series in an online manner from the very beginning, simulating lifelong online learning. We considered a state-of-the art Transformer for LSTF (Informer \cite{informer}), structured convolution-based architectures (FSNet \cite{pham2022learning}), vanilla LSTMs with truncated BPTT (with and without resetting the state when moving from a truncated-BPTT-window to the following one), and RNNs with online computation of the gradients (RTRL, UORO \cite{marschall2020unified}). Some models require a finite-length sequence as input, and in such cases we segmented the time-series using a sliding Window (different sizes, chosen accordingly to previous experiences on this problem \cite{pham2022learning}). We exploited the AdamW optimizer (we also evaluated vanilla GD), and the same architectures and optimal parameter values suggested by the respective authors in their experiences with these data (when available) as starting configurations, and cross-validated new values around them (learning rates, $\gamma$ in FSNet, $label\_len$ param up to window size in Informer).\footnote{In FSNet, Synth data, we considered the second-best sets of parameter values, since the best ones were yielding numerical errors.} In the case of LSTMs and RNNs, we used simpler models, with $8$, $16$, $32$, $128$ or $256$ hidden units, and the following learning rates, $10^{-4}$, $10^{-3}$, $10^{-2}$. 
 We compared results with the ones of a Dummy model which, at time $t$, simply outputs the data received at time $t-1$.
\begin{table*}[!ht]
    \small
    \centering
    \resizebox{0.93\textwidth}{!}{\begin{tabular}{ll|ll|ll|ll|ll|ll|ll}
    \toprule
    \textbf{Data} $\rightarrow$&&\multicolumn{2}{c|}{\bf Synth}&\multicolumn{2}{c|}{\bf ETTh1}&\multicolumn{2}{c|}{\bf ECL}&\multicolumn{2}{c|}{\bf WTH}&\multicolumn{2}{c|}{\bf S-A}&\multicolumn{2}{c}{\bf S-G} \\
    Model $\downarrow$ & Window & {\it {\it MSE}} & {\it MAE} & {\it MSE} & {\it MAE} & {\it MSE} & {\it MAE} & {\it MSE} & {\it MAE} & {\it MSE} & {\it MAE} & {\it MSE} & {\it MAE} \\
    \midrule
\multirow{1}{*}{RNN (RTRL)} & - & $9.2e^{-3}$ & 0.051 & 0.381 & 0.412 & 125.481 & 1.022 & 0.337 & 0.397 & {\it 1.373}$^{\dagger}$ & {\it 0.936}$^{\dagger}$ & {\it 1.297}$^{\dagger}$ & {\it 0.900}$^{\dagger}$\\
\multirow{1}{*}{RNN (UORO)} & - & 2.723 & 0.506 & 0.381 & 0.412 & 277.188 & 1.352 & 3.329 & 0.910 & {\it 1.373}$^{\dagger}$ & {\it 0.936}$^{\dagger}$ & {\it 1.297}$^{\dagger}$ & {\it 0.900}$^{\dagger}$\\
\midrule                                       \multirow{4}{*}{LSTM} & 1 & $4.0e^{-3}$ & 0.011 & {\it 0.301}$^{\dagger}$ & 0.362 & 179.994 & 1.189 & {\it 0.187}$^{\dagger}$ & 0.257 & {\it 1.222}$^{\dagger}$ & {\it 0.884}$^{\dagger}$ & {\it 1.262}$^{\dagger}$ & {\it 0.888}$^{\dagger}$\\
 & 10 & $2.7e^{-3}$ & 0.011 & 0.435 & 0.448 & 140.911 & 1.051 & 0.210 & 0.287 & {\it 1.230}$^{\dagger}$ & {\it 0.886}$^{\dagger}$ & {\it 1.201}$^{\dagger}$ & {\it 0.864}$^{\dagger}$\\
 & 60 & $1.4e^{-3}$ & 0.009 & {\it 0.339}$^{\dagger}$ & 0.392 & 141.825 & 1.053 & 0.212 & 0.288 & {\it 1.216}$^{\dagger}$ & {\it 0.882}$^{\dagger}$ & {\it 1.213}$^{\dagger}$ & {\it 0.868}$^{\dagger}$\\
 & 100 & $1.7e^{-3}$ & 0.010 & {\it 0.290}$^{\dagger}$ & 0.357 & 265.110 & 1.515 & 0.212 & 0.286 & {\it 1.215}$^{\dagger}$ & {\it 0.883}$^{\dagger}$ & {\it 1.217}$^{\dagger}$ & {\it 0.870}$^{\dagger}$\\
\midrule
\multirow{4}{*}{LSTM (No Reset)} & 1 & $2.0e^{-3}$ & 0.009 & {\it 0.347}$^{\dagger}$ & 0.392 & 261.114 & 1.491 & {\it 0.206}$^{\dagger}$ & 0.280 & {\it 1.205}$^{\dagger}$ & {\it 0.879}$^{\dagger}$ & {\it 1.209}$^{\dagger}$ & {\it 0.866}$^{\dagger}$\\
 & 10 & $1.1e^{-3}$ & 0.011 & 0.377 & 0.410 & 256.850 & 1.493 & {\it 0.200}$^{\dagger}$ & 0.273 & {\it 1.226}$^{\dagger}$ & {\it 0.884}$^{\dagger}$ & {\it 1.202}$^{\dagger}$ & {\it 0.864}$^{\dagger}$\\
 & 60 & $1.2e^{-3}$ & 0.008 & {\it 0.316}$^{\dagger}$ & 0.374 & 145.238 & 1.089 & {\it 0.204}$^{\dagger}$ & 0.276 & {\it 1.269}$^{\dagger}$ & {\it 0.901}$^{\dagger}$ & {\it 1.195}$^{\dagger}$ & {\it 0.862}$^{\dagger}$\\
 & 100 & $1.1e^{-3}$ & 0.008 & 0.368 & 0.409 & 271.016 & 1.505 & {\it 0.204}$^{\dagger}$ & 0.278 & {\it 1.229}$^{\dagger}$ & {\it 0.886}$^{\dagger}$ & {\it 1.265}$^{\dagger}$ & {\it 0.888}$^{\dagger}$\\
\midrule
\multirow{4}{*}{FSNet} & 1 & 0.028 & 0.084 & {\it 0.326}$^{\dagger}$ & 0.358 & {\it 3.470}$^{\dagger}$ & 0.546 & {\it 0.182}$^{\dagger}$ & 0.241 & {\it 1.370}$^{\dagger}$ & {\it 0.930}$^{\dagger}$ & 1.419 & 0.938\\
 & 10 & $0.041$ & $0.085$ & {\it 0.265}$^{\dagger}$ & {\it 0.331}$^{\dagger}$ & {\it 3.338}$^{\dagger}$ & 0.552 & {\it 0.181}$^{\dagger}$ & 0.241 & {\it 1.275}$^{\dagger}$ & {\it 0.901}$^{\dagger}$ & {\it 1.311}$^{\dagger}$ & {\it 0.907}$^{\dagger}$\\
 & 60 & $0.054$ & $0.094 $ & {\it 0.260}$^{\dagger}$ & {\it 0.328}$^{\dagger}$ & {\it 3.276}$^{\dagger}$ & 0.549 & {\it 0.180}$^{\dagger}$ & 0.241 & {\it 1.349}$^{\dagger}$ & {\it 0.924}$^{\dagger}$ & {\it 1.310}$^{\dagger}$ & {\it 0.902}$^{\dagger}$\\
 & 100 & $0.046$ & $0.087$ & {\it 0.257}$^{\dagger}$ & {\it 0.326}$^{\dagger}$ & {\it 3.218}$^{\dagger}$ & 0.542 & {\it 0.181}$^{\dagger}$ & 0.243 & {\it 1.337}$^{\dagger}$ & {\it 0.923}$^{\dagger}$ & {\it 1.340}$^{\dagger}$ & {\it 0.919}$^{\dagger}$\\
\midrule
\multirow{4}{*}{Informer} & 1 & $9.7e^{-5}$ & 0.006 & {\it 0.357}$^{\dagger}$ & 0.380 & 309.722 & 1.511 & {\it 0.166}$^{\dagger}$ & 0.224 & {\it 1.204}$^{\dagger}$ & {\it 0.877}$^{\dagger}$ & {\it 1.076}$^{\dagger}$ & {\it 0.821}$^{\dagger}$\\
 & 10 & $3.7e^{-4}$ & 0.010 & {\it 0.287}$^{\dagger}$ & {\it 0.354}$^{\dagger}$ & 311.685 & 1.512 & {\it 0.165}$^{\dagger}$ & 0.225 & {\it 1.116}$^{\dagger}$ & {\it 0.843}$^{\dagger}$ & {\it 1.299}$^{\dagger}$ & {\it 0.899}$^{\dagger}$\\
 & 60 & $1.4e^{-3}$ & 0.021 & {\it 0.286}$^{\dagger}$ & 0.355 & 311.709 & 1.507 & {\it 0.165}$^{\dagger}$ & 0.224 & {\it 1.121}$^{\dagger}$ & {\it 0.843}$^{\dagger}$ & {\it 1.327}$^{\dagger}$ & {\it 0.911}$^{\dagger}$\\
 & 100 & $4.2e^{-4}$ & 0.011 & {\it 0.289}$^{\dagger}$ & 0.357 & 310.627 & 1.515 & {\it 0.165}$^{\dagger}$ & 0.224 & {\it 1.097}$^{\dagger}$ & {\it 0.835}$^{\dagger}$ & {\it 1.327}$^{\dagger}$ & {\it 0.910}$^{\dagger}$\\
\midrule
FSNet (Offline) & 60 & 0.029 & 0.088 & {\it 0.275$^{\dagger}$}& {\it 0.338$^{\dagger}$} & {\it 3.143$^{\dagger}$} & 0.472 & {\it 0.176$^{\dagger}$} & 0.236 & {\it 1.349$^{\dagger}$} & {\it 0.923$^{\dagger}$} & {\it 1.278$^{\dagger}$} & {\it 0.896$^{\dagger}$}\\
Informer (Offline) & 60 & { ${\it 1.4e^{-3}}^{\dagger}$} & 0.019 & {\it 0.321$^{\dagger}$} & 0.372 & 114.962 & 0.930 & {\it 0.148$^{\dagger}$} & 0.198 & {\it 1.093$^{\dagger}$} & {\it 0.832$^{\dagger}$} & {\it 1.046$^{\dagger}$} & {\it 0.808$^{\dagger}$}\\
\midrule
\multirow{1}{*}{Dummy} & - & $3.0e^{-6}$ & $1.6e^{-3}$ & 0.363 & 0.355 & 6.747 & 0.376 & 0.207 & 0.188 & 1.430 & 0.951 & 1.377 & 0.924\\
    \bottomrule
    \end{tabular}}
    \caption{Mean Squared Error (MSE) and Mean Absolute Error (MAE) on the (long) test portion of the time series. See [Pham {\it et al.}, 2022] for a description of the datasets (all but Synth, which is a sinusoidal signal at frequency of 2.8 mHz, with 36000 samples). Results that overcome the Dummy model are marked with $^{\dagger}$. The most recent models were also trained Offline, shuffling data, for comparison.} 
    \label{tab:results}
    \vskip -3mm
\end{table*}
Of course, this must not be intended as an exhaustive comparison of the state-of-the art, but just a tangible experience to show the challenging nature of the discussed learning setting. Results of Table~\ref{tab:results} show that a Dummy model is better than other methods in several cases (in $48.14\%$ of the results), while in the remaining cases (marked with $^{\dagger}$) the gap in terms of performance of the compared models is not so evident, despite their large structural differences. As a matter of fact, RNN and LSTM are way simpler models than FSNet or Informer, and we are also comparing sliding-window-based models against models that never reset their internal state. In general, most of the results are relatively far from the ideal optimum (i.e., zero). To provide another useful reference, we  considered a setup that departs from the point we are making in this section, i.e., the one of \cite{pham2022learning}, where offline-training (multiple epochs, shuffled data, window size $60$) is exploited, before processing the rest of the series as in the already described experiences. The most recent models are compared in such a setting \cite{pham2022learning,informer}, and in the last two rows of Table~\ref{tab:results} we clearly see that the quality of the results increases when compared to the same window-sized models in the upper portion of the table.
All these considerations suggest that there is room for developing new models which are better at capturing regularities on the input stream in an online manner. 
The small size of the validation part of the series, w.r.t. the test one (typical of lifelong learning problems) is another potential issue to address with models that self-adapt their hyper-parameters over time. 


\section{Conclusions}
\label{sec:discussion}
The improvements in long sequence processing of the last few years, both in terms of performance and scalability, are mostly represented by the resurgence of recurrent models with linear recurrence. Interestingly, the recent literature in the context of Transformers and of State-Space Models share the same ingredients, thus confirming the aforementioned trend. We surveyed the most important works, emphasizing how the research direction on lifelong online learning from a stream of data, intended to be an infinite-length sequence, represents a challenge that is still very open and might be faced starting from this renewed interest in recurrent models. 

\appendix
\section*{Acknowledgments}
This work was supported by TAILOR and by HumanE-AI-Net projects funded by EU Horizon 2020 research and innovation programme, GA No 952215 and No 952026, resp.


\bibliographystyle{named}
\bibliography{ijcai24}

\begin{thebibliography}{}

\bibitem[\protect\citeauthoryear{Bengio \bgroup \em et al.\egroup }{1992}]{gori_mori1992}
Yoshua Bengio, Renato de~Mori, and Marco Gori.
\newblock Learning the dynamic nature of speech with back-propagation for sequences.
\newblock {\em Pattern Recognit. Lett.}, 13(5):375--385, 1992.

\bibitem[\protect\citeauthoryear{Betti \bgroup \em et al.\egroup }{2020}]{gori_cal2020}
Alessandro Betti, Marco Gori, and Stefano Melacci.
\newblock Cognitive action laws: The case of visual features.
\newblock {\em {IEEE} Trans. Neural Networks Learn. Syst.}, 31(3):938--949, 2020.

\bibitem[\protect\citeauthoryear{Betti \bgroup \em et al.\egroup }{2022}]{gori_collas2022}
Alessandro Betti, Lapo Faggi, Marco Gori, Matteo Tiezzi, Simone Marullo, Enrico Meloni, and Stefano Melacci.
\newblock Continual learning through hamilton equations.
\newblock In {\em Conference on Lifelong Learning Agents}, volume 199, pages 201--212. {PMLR}, 2022.

\bibitem[\protect\citeauthoryear{Bulatov \bgroup \em et al.\egroup }{2022}]{bulatov2022recurrent}
Aydar Bulatov, Yury Kuratov, and Mikhail Burtsev.
\newblock Recurrent memory transformer.
\newblock {\em Advances in NeurIPS}, 35:11079--11091, 2022.

\bibitem[\protect\citeauthoryear{Dai \bgroup \em et al.\egroup }{2019}]{Dai2019transformerxl}
Zihang Dai, Zhilin Yang, Yiming Yang, William~W. Cohen, Jaime Carbonell, Quoc~V. Le, and Ruslan Salakhutdinov.
\newblock {Transformer-XL: Attentive Language Models Beyond a Fixed-Length Context}.
\newblock In {\em Proceedings of ACL 2019}, 2019.

\bibitem[\protect\citeauthoryear{Elman}{1990}]{elman1990finding}
Jeffrey~L Elman.
\newblock Finding structure in time.
\newblock {\em Cognitive science}, 14(2):179--211, 1990.

\bibitem[\protect\citeauthoryear{Gemini-Team \bgroup \em et al.\egroup }{2023}]{geminiteam2023gemini}
Gemini-Team, Rohan Anil, and Sebastian~Borgeaud et~al.
\newblock Gemini: A family of highly capable multimodal models, 2023.

\bibitem[\protect\citeauthoryear{Gori and Melacci}{2023}]{collectionlessAI}
M.~Gori and S.~Melacci.
\newblock Collectionless artificial intelligence.
\newblock {\em arXiv 2309.06938}, 2023.

\bibitem[\protect\citeauthoryear{Gori \bgroup \em et al.\egroup }{2010}]{goriu_hammer2010}
Marco Gori, Barbara Hammer, Pascal Hitzler, and Guenther Palm.
\newblock Perspectives and challenges for recurrent neural network training.
\newblock {\em Log. J. {IGPL}}, 18(5):617--619, 2010.

\bibitem[\protect\citeauthoryear{Gruslys \bgroup \em et al.\egroup }{2016}]{gruslys2016memory}
Audrunas Gruslys, R{\'e}mi Munos, Ivo Danihelka, Marc Lanctot, and Alex Graves.
\newblock Memory-efficient backpropagation through time.
\newblock {\em Advances in NeurIPS}, 29, 2016.

\bibitem[\protect\citeauthoryear{Gu \bgroup \em et al.\egroup }{2020}]{gu2020hippo}
Albert Gu, Tri Dao, Stefano Ermon, Atri Rudra, and Christopher R{\'e}.
\newblock Hippo: Recurrent memory with optimal polynomial projections.
\newblock {\em Advances in NeurIPS}, 33:1474--1487, 2020.

\bibitem[\protect\citeauthoryear{Gu \bgroup \em et al.\egroup }{2021a}]{gu2021efficiently}
Albert Gu, Karan Goel, and Christopher Re.
\newblock Efficiently modeling long sequences with structured state spaces.
\newblock In {\em ICLR}, 2021.

\bibitem[\protect\citeauthoryear{Gu \bgroup \em et al.\egroup }{2021b}]{gu2021combining}
Albert Gu, Isys Johnson, Karan Goel, Khaled Saab, Tri Dao, Atri Rudra, and Christopher R{\'e}.
\newblock Combining recurrent, convolutional, and continuous-time models with linear state space layers.
\newblock {\em Advances in NeurIPS}, 34:572--585, 2021.

\bibitem[\protect\citeauthoryear{Gu \bgroup \em et al.\egroup }{2022}]{gu2022parameterization}
Albert Gu, Ankit Gupta, Karan Goel, and Christopher R{\'e}.
\newblock On the parameterization and initialization of diagonal state space models.
\newblock {\em arXiv preprint arXiv:2206.11893}, 2022.

\bibitem[\protect\citeauthoryear{Gunasekara \bgroup \em et al.\egroup }{2023}]{streaming}
Nuwan Gunasekara, Bernhard Pfahringer, Heitor~Murilo Gomes, and Albert Bifet.
\newblock Survey on online streaming continual learning.
\newblock In {\em Proc. of IJCAI}, pages 6628--6637, 2023.

\bibitem[\protect\citeauthoryear{Gupta \bgroup \em et al.\egroup }{2022}]{gupta2022diagonal}
Ankit Gupta, Albert Gu, and Jonathan Berant.
\newblock Diagonal state spaces are as effective as structured state spaces.
\newblock In {\em Advances in NeurIPS}, 2022.

\bibitem[\protect\citeauthoryear{Hochreiter and Schmidhuber}{1997}]{hochreiter1997long}
Sepp Hochreiter and J{\"u}rgen Schmidhuber.
\newblock Long short-term memory.
\newblock {\em Neural computation}, 9(8):1735--1780, 1997.

\bibitem[\protect\citeauthoryear{Huang \bgroup \em et al.\egroup }{2022}]{huang2022encoding}
Feiqing Huang, Kexin Lu, CAI Yuxi, Zhen Qin, Yanwen Fang, Guangjian Tian, and Guodong Li.
\newblock Encoding recurrence into transformers.
\newblock In {\em ICLR}, 2022.

\bibitem[\protect\citeauthoryear{Hutchins \bgroup \em et al.\egroup }{2022}]{hutchins2022block}
DeLesley Hutchins, Imanol Schlag, Yuhuai Wu, Ethan Dyer, and Behnam Neyshabur.
\newblock Block-recurrent transformers.
\newblock {\em Advances in NeurIPS}, 35:33248--33261, 2022.

\bibitem[\protect\citeauthoryear{Irie \bgroup \em et al.\egroup }{2023}]{irie2023exploring}
Kazuki Irie, Anand Gopalakrishnan, and J{\"u}rgen Schmidhuber.
\newblock Exploring the promise and limits of real-time recurrent learning.
\newblock {\em arXiv preprint arXiv:2305.19044}, 2023.

\bibitem[\protect\citeauthoryear{Javed \bgroup \em et al.\egroup }{2023}]{javed2023online}
Khurram Javed, Haseeb Shah, Rich Sutton, and Martha White.
\newblock Online real-time recurrent learning using sparse connections and selective learning.
\newblock {\em arXiv preprint arXiv:2302.05326}, 2023.

\bibitem[\protect\citeauthoryear{Kag and Saligrama}{2021}]{kag2021training}
Anil Kag and Venkatesh Saligrama.
\newblock Training recurrent neural networks via forward propagation through time.
\newblock In {\em Int. Conf. on Machine Learning}, pages 5189--5200. PMLR, 2021.

\bibitem[\protect\citeauthoryear{Katharopoulos \bgroup \em et al.\egroup }{2020}]{katharopoulos2020transformers}
A.~Katharopoulos, A.~Vyas, N.~Pappas, and F.~Fleuret.
\newblock Transformers are rnns: Fast autoregressive transformers with linear attention.
\newblock In {\em Int. Conf. on Machine Learning}, pages 5156--5165, 2020.

\bibitem[\protect\citeauthoryear{Li \bgroup \em et al.\egroup }{2018}]{li2018independently}
Shuai Li, Wanqing Li, Chris Cook, Ce~Zhu, and Yanbo Gao.
\newblock Independently recurrent neural network (indrnn): Building a longer and deeper rnn.
\newblock In {\em Proc. of the IEEE Conf. on CVPR}, pages 5457--5466, 2018.

\bibitem[\protect\citeauthoryear{Li \bgroup \em et al.\egroup }{2022}]{li2022approximation}
Zhong Li, Jiequn Han, Weinan E, and Qianxiao Li.
\newblock Approximation and optimization theory for linear continuous-time recurrent neural networks.
\newblock {\em Journal of Machine Learning Research}, 23(1):1997--2081, 2022.

\bibitem[\protect\citeauthoryear{Lipton \bgroup \em et al.\egroup }{2015}]{lipton2015critical}
Z.~C. Lipton, J.~Berkowitz, and C.~Elkan.
\newblock A critical review of recurrent neural networks for sequence learning.
\newblock {\em arXiv:1506.00019}, 2015.

\bibitem[\protect\citeauthoryear{Marschall \bgroup \em et al.\egroup }{2020}]{marschall2020unified}
Owen Marschall, Kyunghyun Cho, and Cristina Savin.
\newblock A unified framework of online learning algorithms for training recurrent neural networks.
\newblock {\em Journal of Machine Learning Research}, 21(1):5320--5353, 2020.

\bibitem[\protect\citeauthoryear{Martin and Cundy}{2018}]{martin2018parallelizing}
Eric Martin and Chris Cundy.
\newblock Parallelizing linear recurrent neural nets over sequence length.
\newblock In {\em ICRL}, 2018.

\bibitem[\protect\citeauthoryear{Massaroli \bgroup \em et al.\egroup }{2023}]{massaroli2023laughing}
Stefano Massaroli, Michael Poli, Daniel~Y Fu, Hermann Kumbong, et~al.
\newblock Laughing hyena distillery: Extracting compact recurrences from convolutions.
\newblock In {\em Advances in NeurIPS}, 2023.

\bibitem[\protect\citeauthoryear{Orvieto \bgroup \em et al.\egroup }{2023}]{orvieto2023resurrecting}
Antonio Orvieto, Samuel~L Smith, Albert Gu, Anushan Fernando, Caglar Gulcehre, Razvan Pascanu, and Soham De.
\newblock Resurrecting recurrent neural networks for long sequences.
\newblock {\em arXiv preprint arXiv:2303.06349}, 2023.

\bibitem[\protect\citeauthoryear{Pascanu \bgroup \em et al.\egroup }{2013}]{pascanu2013difficulty}
Razvan Pascanu, Tomas Mikolov, and Yoshua Bengio.
\newblock On the difficulty of training recurrent neural networks.
\newblock In {\em Int. Conf. on ML}, pages 1310--1318, 2013.

\bibitem[\protect\citeauthoryear{Peng \bgroup \em et al.\egroup }{2021}]{peng2021random}
Hao Peng, Nikolaos Pappas, Dani Yogatama, Roy Schwartz, Noah Smith, and Lingpeng Kong.
\newblock Random feature attention.
\newblock In {\em ICLR}, 2021.

\bibitem[\protect\citeauthoryear{Peng \bgroup \em et al.\egroup }{2023}]{peng2023rwkv}
Bo~Peng, Eric Alcaide, Quentin Anthony, et~al.
\newblock Rwkv: Reinventing rnns for the transformer era.
\newblock {\em arXiv preprint arXiv:2305.13048}, 2023.

\bibitem[\protect\citeauthoryear{Pham \bgroup \em et al.\egroup }{2022}]{pham2022learning}
Quang Pham, Chenghao Liu, Doyen Sahoo, and Steven Hoi.
\newblock Learning fast and slow for online time series forecasting.
\newblock In {\em ICLR}, 2022.

\bibitem[\protect\citeauthoryear{Qin \bgroup \em et al.\egroup }{2022}]{qin2022devil}
Zhen Qin, Xiaodong Han, Weixuan Sun, Dongxu Li, Lingpeng Kong, Nick Barnes, and Yiran Zhong.
\newblock The devil in linear transformer.
\newblock In {\em Proceedings of EMNLP}, pages 7025--7041, 2022.

\bibitem[\protect\citeauthoryear{Qin \bgroup \em et al.\egroup }{2023}]{qin2023hierarchically}
Zhen Qin, Songlin Yang, and Yiran Zhong.
\newblock Hierarchically gated recurrent neural network for sequence modeling.
\newblock In {\em Advances in NeurIPS}, 2023.

\bibitem[\protect\citeauthoryear{Rubanova \bgroup \em et al.\egroup }{2019}]{rubanova2019latent}
Yulia Rubanova, Ricky~TQ Chen, and David~K Duvenaud.
\newblock Latent ordinary differential equations for irregularly-sampled time series.
\newblock {\em Advances in NeurIPS}, 32, 2019.

\bibitem[\protect\citeauthoryear{Rumelhart \bgroup \em et al.\egroup }{1985}]{rumelhart1985learning}
David~E Rumelhart, Geoffrey~E Hinton, Ronald~J Williams, et~al.
\newblock Learning internal representations by error propagation, 1985.

\bibitem[\protect\citeauthoryear{Salehinejad \bgroup \em et al.\egroup }{2017}]{salehinejad2017recent}
Hojjat Salehinejad, Sharan Sankar, Joseph Barfett, Errol Colak, and Shahrokh Valaee.
\newblock Recent advances in recurrent neural networks.
\newblock {\em arXiv preprint arXiv:1801.01078}, 2017.

\bibitem[\protect\citeauthoryear{Sch{\"a}fer and Zimmermann}{2006}]{schafer2006recurrent}
Anton Sch{\"a}fer and Hans~Georg Zimmermann.
\newblock Recurrent neural networks are universal approximators.
\newblock In {\em Proceedings of ICANN}, pages 632--640. Springer, 2006.

\bibitem[\protect\citeauthoryear{Schlag \bgroup \em et al.\egroup }{2021}]{schlag2021linear}
Imanol Schlag, Kazuki Irie, and J{\"u}rgen Schmidhuber.
\newblock Linear transformers are secretly fast weight programmers.
\newblock In {\em ICML}, pages 9355--9366. PMLR, 2021.

\bibitem[\protect\citeauthoryear{Schmidhuber}{1992}]{schmidhuber1992learning}
J{\"u}rgen Schmidhuber.
\newblock Learning to control fast-weight memories: An alternative to dynamic recurrent networks.
\newblock {\em Neural Comp.}, 4(1):131--139, 1992.

\bibitem[\protect\citeauthoryear{Selva \bgroup \em et al.\egroup }{2023}]{selva2023video}
Javier Selva, Anders~S Johansen, Sergio Escalera, Kamal Nasrollahi, Thomas~B Moeslund, and Albert Clap{\'e}s.
\newblock Video transformers: A survey.
\newblock {\em IEEE Transactions on Pattern Analysis and Machine Intelligence}, 2023.

\bibitem[\protect\citeauthoryear{Siegelmann}{2012}]{siegelmann2012neural}
Hava~T Siegelmann.
\newblock {\em Neural networks and analog computation: beyond the Turing limit}.
\newblock Springer Science \& Business Media, 2012.

\bibitem[\protect\citeauthoryear{Smith \bgroup \em et al.\egroup }{2022}]{smith2022simplified}
Jimmy~TH Smith, Andrew Warrington, and Scott~W Linderman.
\newblock Simplified state space layers for sequence modeling.
\newblock {\em preprint arXiv:2208.04933}, 2022.

\bibitem[\protect\citeauthoryear{Sun \bgroup \em et al.\egroup }{2023}]{sun2023retentive}
Y.~Sun, L.~Dong, S.~Huang, S.~Ma, Y.~Xia, J.~Xue, J.~Wang, and F.~Wei.
\newblock Retentive network: A successor to transformer for large language models.
\newblock {\em arXiv:2307.08621}, 2023.

\bibitem[\protect\citeauthoryear{Sutskever \bgroup \em et al.\egroup }{2014}]{sutskever2014sequence}
Ilya Sutskever, Oriol Vinyals, and Quoc~V Le.
\newblock Sequence to sequence learning with neural networks.
\newblock {\em arXiv preprint arXiv:1409.3215}, 2014.

\bibitem[\protect\citeauthoryear{Tay \bgroup \em et al.\egroup }{2020}]{tay2020long}
Yi~Tay, Mostafa Dehghani, Samira Abnar, et~al.
\newblock Long range arena: A benchmark for efficient transformers.
\newblock {\em arXiv preprint arXiv:2011.04006}, 2020.

\bibitem[\protect\citeauthoryear{Tay \bgroup \em et al.\egroup }{2023}]{tayefficient}
Yi~Tay, Mostafa Dehghani, Dara Bahri, and Donald Metzler.
\newblock Efficient transformers: {A} survey.
\newblock {\em {ACM} Comput. Surv.}, 55(6):109:1--109:28, 2023.

\bibitem[\protect\citeauthoryear{Vaswani \bgroup \em et al.\egroup }{2017}]{vaswani2017attention}
Ashish Vaswani, Noam Shazeer, Niki Parmar, Jakob Uszkoreit, Llion Jones, Aidan~N Gomez, {\L}ukasz Kaiser, and Illia Polosukhin.
\newblock Attention is all you need.
\newblock {\em Advances in NeurIPS}, 30, 2017.

\bibitem[\protect\citeauthoryear{Voelker \bgroup \em et al.\egroup }{2019}]{voelker2019}
Aaron Voelker, Ivana Kaji\'{c}, and Chris Eliasmith.
\newblock Legendre memory units: Continuous-time representation in recurrent neural networks.
\newblock In {\em Advances in NeurIPS}, volume~32, 2019.

\bibitem[\protect\citeauthoryear{Werbos}{1990}]{werbos1990backpropagation}
Paul~J Werbos.
\newblock Backpropagation through time: what it does and how to do it.
\newblock {\em Proceedings of the IEEE}, 78(10):1550--1560, 1990.

\bibitem[\protect\citeauthoryear{Williams and Peng}{1990}]{williams1990efficient}
Ronald~J Williams and Jing Peng.
\newblock An efficient gradient-based algorithm for on-line training of recurrent network trajectories.
\newblock {\em Neural computation}, 2(4):490--501, 1990.

\bibitem[\protect\citeauthoryear{Xu \bgroup \em et al.\egroup }{2015}]{xu2015show}
Kelvin Xu, Jimmy Ba, Ryan Kiros, et~al.
\newblock Show, attend and tell: Neural image caption generation with visual attention.
\newblock In {\em ICML}, pages 2048--2057, 2015.

\bibitem[\protect\citeauthoryear{Yu \bgroup \em et al.\egroup }{2019}]{yu2019review}
Yong Yu, Xiaosheng Si, Changhua Hu, and Jianxun Zhang.
\newblock A review of recurrent neural networks: Lstm cells and network architectures.
\newblock {\em Neural computation}, 31(7):1235--1270, 2019.

\bibitem[\protect\citeauthoryear{Zhai \bgroup \em et al.\egroup }{2021}]{zhai2021attention}
Shuangfei Zhai, Walter Talbott, Nitish Srivastava, Chen Huang, Hanlin Goh, Ruixiang Zhang, and Josh Susskind.
\newblock An attention free transformer.
\newblock {\em arXiv preprint arXiv:2105.14103}, 2021.

\bibitem[\protect\citeauthoryear{Zhou \bgroup \em et al.\egroup }{2021}]{informer}
Haoyi Zhou, Shanghang Zhang, Jieqi Peng, et~al.
\newblock Informer: Beyond efficient transformer for long sequence time-series forecasting.
\newblock {\em Proceedings of the AAAI Conf. on AI}, 35(12):11106--11115, May 2021.

\bibitem[\protect\citeauthoryear{Zucchet \bgroup \em et al.\egroup }{2023}]{zucchet2023online}
Nicolas Zucchet, Robert Meier, Simon Schug, Asier Mujika, and Jo{\~a}o Sacramento.
\newblock Online learning of long range dependencies.
\newblock {\em NeurIPS}, 2023.

\end{thebibliography}

\end{document}